\documentclass{article}
\usepackage{preprint,times}

\usepackage[T1]{fontenc}
\usepackage[utf8]{inputenc}
\usepackage{microtype}
\usepackage{amsmath,amssymb,amsthm}
\usepackage{booktabs}
\usepackage{multirow}
\usepackage{graphicx}
\usepackage{xcolor}
\usepackage{algorithm}
\usepackage{algpseudocode}
\usepackage{hyperref}
\usepackage{url}

\newtheorem{proposition}{Proposition}
\newtheorem{definition}{Definition}

\newcommand{\sig}{\sigma}

\title{Measured Coupling for Parallel Decoding\\in Diffusion Language Models}
\author{Tal Oved \\
Tel Aviv University \And
Lior Wolf \\
Tel Aviv University}

\begin{document}
\maketitle

\begin{abstract}
Diffusion language models can write many tokens per step, but must choose which tokens are safe to commit together.
Mean-field decoding penalizes pairs of positions whose predictions look alike.
We replace this proxy with the coupling itself: how much committing one token changes the model's confidence at another.
Cheap probes inside the main forward pass measure it, with no training or extra model, and it plugs into the same mean-field rule.
On three models and four benchmarks, our method needs far fewer steps than mean-field decoding and is faster in most settings, also with a key-value cache, at an average accuracy cost of 1.1 points.
Shuffling the couplings or replacing them with a uniform bonus loses this gain.
\end{abstract}

\section{Introduction}

Large language models \citep{achiam2023gpt4,bai2023qwen,comanici2025gemini,liu2024deepseek} write text one token at a time.
Each new token needs a full forward pass, so the time to answer grows with the length of the answer.
Discrete diffusion language models \citep{austin2021d3pm,li2022diffusionlm,nie2025llada,zhu2025llada15,ye2025dream} offer a parallel alternative.
They treat generation as iterative denoising: the answer starts fully masked, and every step can fill in several positions at once.
Recent models scale this idea to strong, large systems \citep{bie2025llada20,bie2026llada21}, which makes fast parallel decoding a practical goal.

The hard part is deciding which tokens to commit together.
A diffusion model predicts every masked position independently, from the same context.
Two tokens can each look certain on their own and still form a wrong pair, because neither was predicted with the other in view \citep{demask2026,luo2026dawn,zhang2026order}.
Confidence thresholds commit only positions whose top prediction is very likely \citep{wu2025fastdllm}.
Dependency-aware decoders estimate which positions interact, from attention maps \citep{luo2026dawn,dapd2026} or with a small trained predictor \citep{demask2026}.
Mean-field parallel decoding \citep{zoabi2026meanfield} takes a training-free view based on an energy model.
Each masked position gets a commit score, a pairwise term penalizes committing two positions whose predictive distributions are similar, measured by the Jensen-Shannon divergence, and a few mean-field updates decide which positions to commit.

\begin{figure}[t]
\centering
\includegraphics[width=\linewidth]{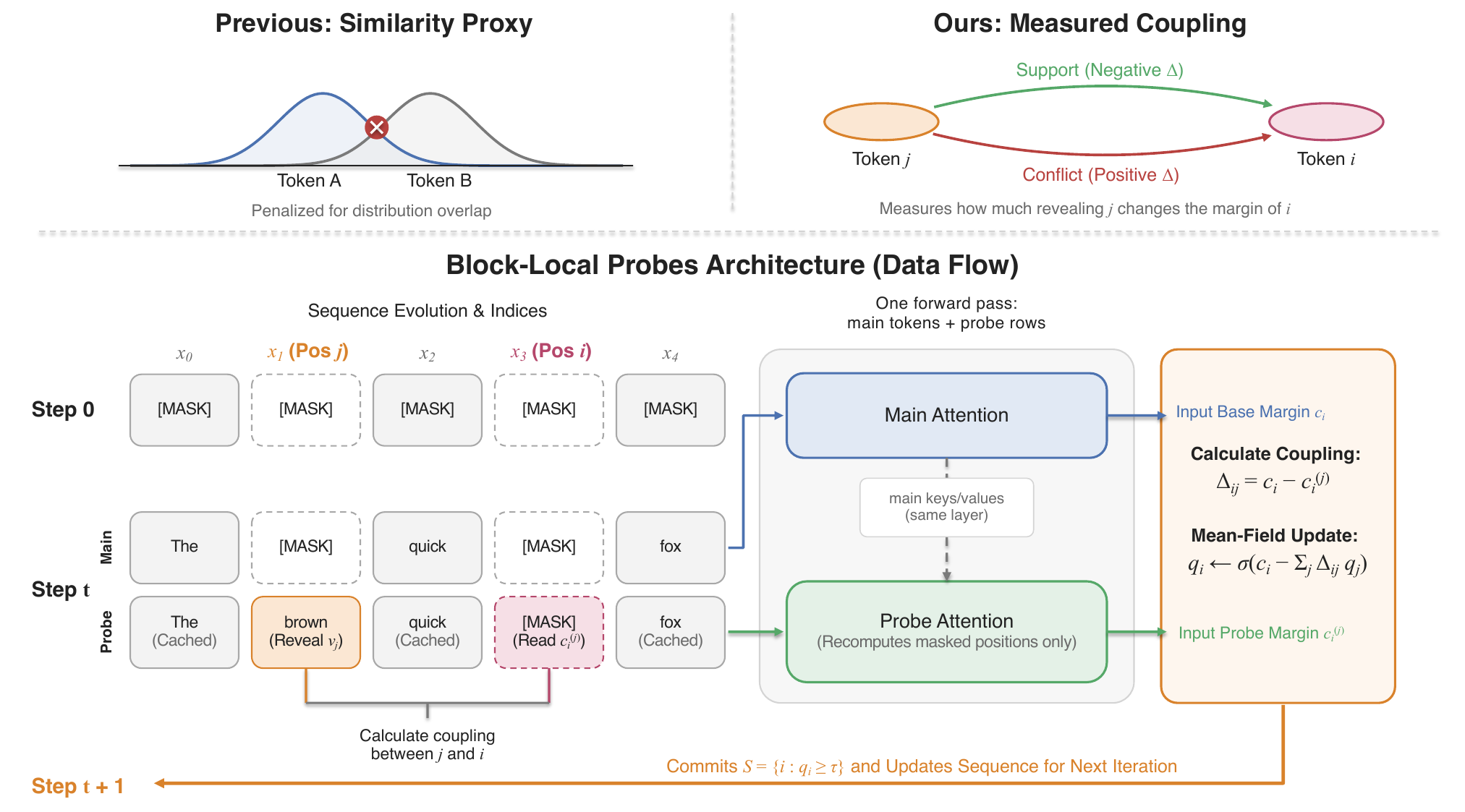}
\caption{Overview of measured coupling. Top: mean-field decoding penalizes two positions whose predicted distributions overlap (left). We instead measure how much revealing token $j$ changes the margin of token $i$, which shows support as well as conflict (right). Bottom: one decoding step. The main tokens and $K$ probe rows, one for each of the most confident candidates, run in a single forward pass. Each probe row reveals one candidate and recomputes only the masked positions, reading the main keys and values of the same layer everywhere else. The resulting margins give the coupling $\Delta_{ij} = c_i - c_i^{(j)}$, which enters the mean-field update, and positions with $q_i \ge \tau$ are committed for the next step.}
\label{fig:overview}
\end{figure}

In this work, we tackle the question of what pairwise quantity should this energy use.
The quantity that matters is the effect of committing token $j$ on the confidence of position $i$.
We call it the \emph{coupling} $\Delta_{ij}$ (Figure~\ref{fig:overview}).
It can be positive, when $j$ makes $i$ less certain, or negative, when $j$ supports $i$.
Two positions with similar distributions may not affect each other at all, and two positions with different distributions may depend strongly on each other, for example the digits of one number.

The coupling can be measured with the model itself: reveal $j$, run the model, and read the new confidence at $i$.
Done naively, this costs one forward pass per candidate.
We make it cheap.
A probe only needs to recompute the current block.
The rest of the sequence reuses the keys and values from the step's main forward pass.
This approximation is accurate (Pearson $0.96$ to $0.97$ with exact probes) and reduces the work of a probe from the whole sequence to the masked positions of one block.
We then run the probes inside the main forward pass of each step, so the model weights are read once for both.

\newpage

Our contributions are:
\begin{itemize}
\item \textbf{Measured coupling.} We define the pairwise term of parallel decoding as the change in log-margin caused by committing another token, including support as well as conflict.
\item \textbf{Fused block-local probes.} A probe recomputes only the masked positions of the current block against the keys and values of the main pass, and all probes run inside that same pass. This makes measuring the coupling cheap enough to pay off in wall-clock time.
\item \textbf{A reading of the mean-field rule.} As an interpretation, under an additive model of conditional margins the mean-field fixed point with measured coupling equals the sigmoid of each token's expected margin given the others (Proposition~\ref{prop:expected}), and a standard contraction condition guarantees a unique fixed point (Proposition~\ref{prop:contraction}).
\item \textbf{Results.} On LLaDA-8B, LLaDA-1.5, and Dream, measured coupling needs the fewest steps on every benchmark and has the shortest time per answer in most of them. Its accuracy is 1.1 points below mean-field decoding on average and ranges from 6.1 points lower to 1.8 points higher; the drop is significant on MATH for two of the three models, and the largest drops are on Dream HumanEval and LLaDA-1.5 MBPP.
\end{itemize}

\section{Background}

\paragraph{Decoding a masked diffusion model.}
The sequence $x$ holds a prompt followed by masked positions.
Decoding fills the answer block by block.
At each step the model reads $x$ and returns a distribution $\pi_i$ for every masked position $i$ in the current block.
Let $U$ be these positions.
For each $i \in U$ we take the most likely token $v_i = \arg\max \pi_i$ and its log-margin
\begin{equation}
c_i = \log \pi_i(v_i) - \max_{v \neq v_i} \log \pi_i(v).
\end{equation}
The decoder commits a set $S \subseteq U$, writes $v_S$ into $x$, and repeats until the block is full.
The number of decoding steps, called function evaluations (NFE), sets the cost.

\paragraph{Mean-field parallel decoding.}
\citet{zoabi2026meanfield} attach a binary variable $s_i$ to each $i \in U$, with $s_i = 1$ meaning commit.
They define the energy
\begin{equation}
\label{eq:energy}
E(s) = -\sum_i c_i s_i + \sum_{i < j} D_{ij}\, s_i s_j ,
\end{equation}
where $D_{ij}$ is one minus the normalized Jensen-Shannon divergence between $\pi_i$ and $\pi_j$.
A factorized distribution $q$ with $q_i = P(s_i = 1)$ gives the mean-field update
\begin{equation}
\label{eq:mf}
q_i \leftarrow \sig\Big(c_i - \sum_{j \neq i} D_{ij} q_j\Big),
\end{equation}
run for two iterations from $q = \sig(c)$.
Positions with $q_i \ge \tau$ are committed.
This rule is training-free and fast, and it is the method we build on.

\section{Measured Coupling}

\subsection{The coupling}

\begin{definition}[Coupling]
For positions $i, j \in U$, let $c_i^{(j)}$ be the log-margin of the same token $v_i$ at position $i$ after revealing $x_j = v_j$.
The coupling of $j$ on $i$ is
\begin{equation}
\Delta_{ij} = c_i - c_i^{(j)} .
\end{equation}
\end{definition}

A positive $\Delta_{ij}$ means that committing $j$ lowers the confidence at $i$, so committing both together is risky.
A negative $\Delta_{ij}$ means that $j$ supports $i$: once $j$ is known, $i$ becomes more certain.
The coupling is directional, since the effect of $j$ on $i$ can differ from the effect of $i$ on $j$.
We keep its sign.
Support is what allows many tokens to be committed in one step.

We plug the coupling into the same update as \eqref{eq:mf}:
\begin{equation}
\label{eq:mfdelta}
q_i \leftarrow \sig\Big(c_i - \sum_{j \neq i} \Delta_{ij} q_j\Big).
\end{equation}

\subsection{What the update computes}

The next result gives a simple reading of this update.
Write $c_i(x_S)$ for the log-margin of $v_i$ at position $i$ after revealing the tokens $v_S$.

\begin{proposition}[Expected margin]
\label{prop:expected}
Assume that conditional margins are additive in the revealed tokens,
\begin{equation}
\label{eq:additive}
c_i(x_S) = c_i - \sum_{j \in S} \Delta_{ij} \quad \text{for all } S \subseteq U \setminus \{i\}.
\end{equation}
Let the other positions be committed with probabilities $\Pr(s_j = 1) = q_j$, independently or not.
Then the expected margin of $i$ is
\begin{equation}
\mathbb{E}_q\big[c_i(x_S)\big] = c_i - \sum_{j \neq i} \Delta_{ij} q_j ,
\end{equation}
and every fixed point of \eqref{eq:mfdelta} satisfies $q_i = \sig\big(\mathbb{E}_q[c_i(x_S)]\big)$ for all $i$.
\end{proposition}

\begin{proof}
Under \eqref{eq:additive}, $c_i(x_S) = c_i - \sum_{j \neq i} \Delta_{ij} s_j$.
Taking expectations and using $\mathbb{E}[s_j] = q_j$ gives the first claim, by linearity of expectation.
A fixed point of \eqref{eq:mfdelta} satisfies $q_i = \sig(c_i - \sum_{j} \Delta_{ij} q_j)$, which is the sigmoid of this expectation.
\end{proof}

In words, each token's commit probability becomes a function of how confident it is expected to be once the other likely commits are in place.
With $\Delta$ in place of $D$, the mean-field rule of \citet{zoabi2026meanfield} gains this direct reading.
The similarity matrix $D$ can be seen as a cheap stand-in for $\Delta$.

The proposition interprets the update; it does not show that the committed set is safe.
Since $\Delta$ is not symmetric, the update is a fixed-point rule with the mean-field form, and no longer solves a pairwise energy.
The value $\sig(c_i)$ compares only the top two tokens, so it is not a calibrated probability of being correct, and $\sig(\mathbb{E}[c_i])$ differs from $\mathbb{E}[\sig(c_i)]$.
The proposition also averages over random commits, while the decoder commits every position above $\tau$ after two iterations.
Assumption \eqref{eq:additive} is a first-order model, and our probes give estimates of each $\Delta_{ij}$ (Section~\ref{sec:probes}), so additivity and estimation error both enter the final decision.
We test additivity directly in Appendix~\ref{app:additivity}. For the sets the decoder commits, the additive prediction and the jointly measured margin fall on the same side of $\tau$ for 90\% to 93\% of the remaining positions, and for 95\% to 99\% when two or three tokens are committed. The sum overstates both support and conflict, and more so for large sets.

\subsection{When the update has a unique answer}
\label{sec:contraction}

\begin{proposition}[Contraction]
\label{prop:contraction}
Let $F(q)_i = \sig\big(c_i - \sum_{j \neq i} \Delta_{ij} q_j\big)$ and $\|\Delta\|_\infty = \max_i \sum_{j \neq i} |\Delta_{ij}|$.
If $\|\Delta\|_\infty < 4$, then $F$ is a contraction in the $\ell_\infty$ norm.
It has a unique fixed point, and the iteration $q \leftarrow F(q)$ converges to it geometrically from any start.
\end{proposition}

\begin{proof}
The sigmoid satisfies $|\sig'(z)| \le 1/4$.
By the mean value theorem, $|F(q)_i - F(q')_i| \le \tfrac14 \sum_{j} |\Delta_{ij}|\,|q_j - q'_j| \le \tfrac14 \|\Delta\|_\infty \|q - q'\|_\infty$.
The Banach fixed point theorem completes the proof.
\end{proof}

When couplings are large, synchronous updates can swing between states.
A simple example is two strongly conflicting tokens that rise together, fall together, and rise again.
This explains why the number of mean-field iterations matters in practice \citep{zoabi2026meanfield}.
Like mean-field decoding, we run the update for two iterations at every decoding step.
In practice the measured couplings are strong, and the contraction condition holds in about a quarter of the decoding steps.
The condition is therefore sufficient and rarely met in practice.
Running the update for 100 iterations changes the commit decision in about one step in seven. When the update oscillates, two iterations need not be close to a fixed point.

\subsection{Cheap block-local probes}
\label{sec:probes}

Measuring $\Delta_{ij}$ directly needs one forward pass per candidate $j$, over the full sequence.
We reduce this cost with the key-value cache.
The main forward pass of a step already computes keys and values for every position.
A probe for candidate $j$ writes $v_j$ at position $j$ and recomputes only the positions of the current block that are still masked.
All other positions, inside or outside the block, reuse the keys and values of the main pass.
The probes for all candidates run as one batch.

This is the same approximation that Fast-dLLM uses for its dual cache \citep{wu2025fastdllm}.
Here it is applied to probes.
It treats the representations of all other positions as fixed while $j$ is revealed.
Section~\ref{sec:fidelity} shows that the resulting couplings agree closely with exact probes.

\paragraph{Probes inside the main pass.}
A separate probe call costs a fixed amount, mostly for reading the model weights, even when it holds few tokens.
We remove this cost by running the probes inside the main forward pass of the next step.
All token-wise layers (normalization, projections, MLP, and output head) process the main tokens and the probe tokens together, so each weight is read once.
Only attention is split.
The main tokens use the usual bidirectional attention.
Each probe row attends to its own keys at the masked positions and to the main keys and values of the same layer everywhere else, including the block positions that are already committed.
Since the main logits of a step are not known before the pass, the probes use the proposals $v_j$ and the confidence ranking from the previous step, including the predictions for the next block at a block boundary.
A probe whose token has changed since then is dropped.
At the first step of an answer, the probes run in one extra call after the main pass, so an answer takes NFE${}+1$ calls.
We probe the $K$ most confident candidates, which carry most of the weight $q_j$ in the mean-field sum, and set $\Delta_{ij} = 0$ for the others.

\begin{algorithm}[t]
\caption{One decoding step with measured coupling}
\label{alg:step}
\begin{algorithmic}
\Require Sequence $x$, masked positions $\mathcal{U}_t$ of the current block, proposals $v_j$ and margins $c_j$ from the previous step, parameters $\tau$, $R$, $K$
\Ensure Commit set $\mathcal{C}_t$
\State \textit{// 1. Probe construction}
\State $\mathcal{P} \gets$ the $K$ positions $j \in \mathcal{U}_t$ with the largest $c_j$
\State $x^{(j)} \gets$ the tokens at $\mathcal{U}_t$ with $x_j := v_j$, for all $j \in \mathcal{P}$
\State \textit{// 2. One forward pass for $x$ and all probes}
\State $\{\ell_i\},\ \{\ell_i^{(j)}\} \gets f_\theta\big(x, \{x^{(j)}\}_{j \in \mathcal{P}}\big)$ \Comment{probes reuse the main keys and values outside $\mathcal{U}_t$}
\State \textit{// 3. Local confidence}
\State $\pi_{t,i} \gets \mathrm{softmax}(\ell_i)$, \ $v_i \gets \arg\max_v \pi_{t,i}(v)$ for all $i \in \mathcal{U}_t$
\State $c_i \gets \log \pi_{t,i}(v_i) - \max_{v \neq v_i} \log \pi_{t,i}(v)$, and $c_i^{(j)}$ from $\ell_i^{(j)}$ with the same token $v_i$
\State \textit{// 4. Measured coupling}
\State $\Delta_{ij} \gets c_i - c_i^{(j)}$ for $j \in \mathcal{P}$ whose probed token is still $v_j$, and $\Delta_{ij} \gets 0$ otherwise; $\Delta_{ii} \gets 0$
\State \textit{// 5. Mean-field relaxation}
\State $q^{(0)} \gets \sig(c)$
\For{$r = 0$ to $R - 1$}
\State $q^{(r+1)} \gets \sig\big(c - \Delta q^{(r)}\big)$
\EndFor
\State \textit{// 6. Thresholding}
\State $\mathcal{C}_t \gets \{i \in \mathcal{U}_t : q_i^{(R)} \ge \tau\}$; if it is empty, $\mathcal{C}_t \gets \{\arg\max_i q_i^{(R)}\}$
\State \Return $\mathcal{C}_t$
\end{algorithmic}
\end{algorithm}

Algorithm~\ref{alg:step} summarizes one step.
The method adds no parameters and needs no training data. Code, including the fused probe kernels, exact configurations, evaluation and timing scripts, and per-example outputs, is available at \url{https://github.com/TalOved1/measured-coupling}.

\section{Experiments}

\subsection{Setup}

We use LLaDA-8B-Instruct \citep{nie2025llada} and the DAWN code base \citep{luo2026dawn}, which the mean-field paper also uses.
We also evaluate LLaDA-1.5 \citep{zhu2025llada15} with the same protocol, and Dream-v0-Instruct-7B \citep{ye2025dream}, a model from a different family.
We evaluate on GSM8K \citep{cobbe2021gsm8k} (1{,}319 problems, 5-shot), MATH \citep{hendrycks2021math} (5{,}000 problems, 4-shot with the Minerva prompts \citep{lewkowycz2022minerva}), HumanEval \citep{chen2021codex} (164 problems, 0-shot), and MBPP \citep{austin2021mbpp} (500 problems, 3-shot).
The generation length is 256 tokens.
MATH answers are checked for mathematical equivalence with math-verify, and HumanEval answers are post-processed with the sanitizer from the DAWN code base.

We compare against one-token-per-step decoding, Fast-dLLM with threshold $0.9$ \citep{wu2025fastdllm}, DAWN \citep{luo2026dawn}, and mean-field decoding with the published LLaDA setting (block 20, $\tau = 0.85$, two iterations) \citep{zoabi2026meanfield}.
Our method uses the same block size (20) and number of iterations (two) as mean-field decoding.
All methods are greedy.
Section~\ref{sec:selection} explains how we chose the remaining settings of our method.

We report accuracy, the average number of function evaluations (NFE) per answer, and the step reduction over one-token-per-step decoding ($256 / \text{NFE}$).
Throughput (tokens per second, TPS) depends on the GPU, so we measure it for all methods in the same job on the same A100 GPU.
We compare accuracy with mean-field decoding on the same questions: we report the paired difference with a 95\% confidence interval (paired Wald interval), and use a paired exact McNemar test where we call a difference significant. When an interval includes zero, the data cannot tell which method is more accurate; it does not show that they are equally accurate.

\subsection{How we chose the settings}
\label{sec:selection}

Our method has three settings: the threshold $\tau$, the number of probes $K$, and whether the probes run in a separate call or inside the main pass.
We follow the protocol of mean-field decoding \citep{zoabi2026meanfield}: we choose the settings on the first 100 questions of the GSM8K training split, separately for each model family.
The candidate with the highest held-out accuracy wins, and ties go to the candidate with fewer steps.
This selects $K = 12$ probes inside the main pass with $\tau = 0.9$ for LLaDA-8B-Instruct, which LLaDA-1.5 shares, and $K = 20$ with $\tau = 0.95$ for Dream.
As a check on code, we also scored the candidates on the validation split of MBPP (90 problems), which does not overlap with the MBPP or HumanEval test sets; on LLaDA the two lists overlap but differ, since the candidates changed during development.
On Dream it selects the same setting.
On LLaDA its best threshold is slightly higher, $\tau = 0.95$, two problems ahead of $\tau = 0.9$.
Tables~\ref{tab:heldout} and~\ref{tab:heldout_dream} in the appendix list every candidate scored on each set.

All baselines use their published settings.
Mean-field decoding uses the LLaDA setting of \citet{zoabi2026meanfield} (block 20, $\tau = 0.85$, two iterations), and Fast-dLLM and DAWN use the settings of the DAWN code base.
On Dream, mean-field decoding uses the default Dream setting of \citet{zoabi2026meanfield} (block 20, $\tau = 0.95$, two iterations), and Fast-dLLM and DAWN use the Dream settings of the DAWN code base.
Their test results are therefore not tuned on the test sets either.

During development we also evaluated other variants on the test sets, and Section~\ref{sec:analysis} reports several of them as analysis.
The main table uses only the settings chosen by the rule above.
The candidate list grew once during the project: we added the probe variant that recomputes only the masked positions, and applied the same rule again to the full list.
All reported results use the setting selected from the full list.

\subsection{Main results}

\begin{table}[t]
\centering
\caption{Results on LLaDA-8B-Instruct. Step reduction is $256/\text{NFE}$, relative to one-token-per-step decoding, and is not a runtime speedup. TPS (tokens per second) is measured for all methods of a benchmark in one job on the same GPU (GSM8K on its first 200 questions, MATH on 105 problems with 15 per subject, HumanEval and MBPP on the full test sets). Time is the wall-clock time per answer in seconds, from the same job. Bold marks the best value in each efficiency column. Ours is measured coupling, MF is mean-field decoding, and one token is one-token-per-step decoding. The last column is the accuracy difference from mean-field decoding in points, with a paired 95\% confidence interval.}
\label{tab:main}
\small
\setlength{\tabcolsep}{3.5pt}
\begin{tabular}{llcccccc}
\toprule
Bench. & Method & Acc.\ $\uparrow$ & NFE $\downarrow$ & Step red. $\uparrow$ & TPS $\uparrow$ & Time (s) $\downarrow$ & $\Delta$ vs.\ MF [95\% CI] \\
\midrule
\multirow{5}{*}{GSM8K}
 & One token & 77.94 & 256.0 & 1.0$\times$ & 7.2 & 32.42 & \\
 & Fast-dLLM & 78.17 & 78.5 & 3.3$\times$ & 23.2 & 10.00 & \\
 & DAWN & 77.79 & 55.8 & 4.6$\times$ & 31.7 & 7.29 & \\
 & MF & 76.57 & 65.4 & 3.9$\times$ & 28.6 & 8.10 & \\
 & Ours & 76.27 & \textbf{40.1} & \textbf{6.4$\times$} & \textbf{40.3} & \textbf{5.86} & $-0.3$ [$-2.5$, $+1.9$] \\
\midrule
\multirow{5}{*}{MATH}
 & One token & 33.82 & 256.0 & 1.0$\times$ & 9.4 & 26.21 & \\
 & Fast-dLLM & 33.36 & 97.0 & 2.6$\times$ & 24.0 & 10.32 & \\
 & DAWN & 32.30 & 72.6 & 3.5$\times$ & 30.8 & 8.01 & \\
 & MF & 32.02 & 77.2 & 3.3$\times$ & 30.8 & 8.18 & \\
 & Ours & 31.18 & \textbf{47.9} & \textbf{5.3$\times$} & \textbf{41.5} & \textbf{6.05} & $-0.8$ [$-1.8$, $+0.2$] \\
\midrule
\multirow{5}{*}{HumanEval}
 & One token & 41.46 & 256.0 & 1.0$\times$ & 19.8 & 12.27 & \\
 & Fast-dLLM & 42.68 & 78.9 & 3.2$\times$ & 64.5 & 3.76 & \\
 & DAWN & 37.80 & 61.6 & 4.2$\times$ & 80.6 & 2.97 & \\
 & MF & 37.20 & 61.7 & 4.1$\times$ & 80.4 & 2.93 & \\
 & Ours & 39.02 & \textbf{40.3} & \textbf{6.4$\times$} & \textbf{83.7} & \textbf{2.77} & $+1.8$ [$-4.1$, $+7.8$] \\
\midrule
\multirow{5}{*}{MBPP}
 & One token & 30.00 & 256.0 & 1.0$\times$ & 6.4 & 26.92 & \\
 & Fast-dLLM & 30.20 & 70.6 & 3.6$\times$ & 23.4 & 7.41 & \\
 & DAWN & 31.00 & 50.5 & 5.1$\times$ & 30.7 & 5.42 & \\
 & MF & 30.20 & 51.6 & 5.0$\times$ & 28.6 & 5.31 & \\
 & Ours & 29.00 & \textbf{32.3} & \textbf{7.9$\times$} & \textbf{31.7} & \textbf{4.00} & $-1.2$ [$-4.9$, $+2.5$] \\
\bottomrule
\end{tabular}
\end{table}

Table~\ref{tab:main} shows the main results.
Measured coupling needs the fewest steps on every benchmark, more than a third fewer than mean-field decoding.
Its accuracy differs from mean-field decoding by $-1.2$ to $+1.8$ points, and every 95\% confidence interval of the paired difference includes zero.

The step savings carry over to wall-clock time.
On the same GPU, our method has the highest throughput of all methods on every benchmark, most of all on GSM8K, whose long prompts make the main pass dominate a step.
Appendix~\ref{app:latency} compares the methods at matched accuracy, with every threshold timed on the same GPU. On GSM8K and MBPP our method reaches a given accuracy in less time than mean-field decoding where both reach it, and on HumanEval the two curves cross.
With the dual key-value cache of Fast-dLLM for all methods, our method at block 32 is the fastest on GSM8K and MBPP (Appendix~\ref{app:cache}).

Figure~\ref{fig:curves} in Appendix~\ref{app:curves} shows the full trade-off between accuracy and steps, sweeping the threshold of each method.
On GSM8K, HumanEval, and MBPP our curve lies to the left of the mean-field curve: it reaches a given accuracy with far fewer steps, at accuracies up to 76.7\% on GSM8K.
On MBPP it also rises above the best mean-field point.

Table~\ref{tab:llada15} repeats the comparison on LLaDA-1.5, with the setting chosen on LLaDA-8B-Instruct and no new tuning.
Our method again needs the fewest steps on every benchmark, 1.3 to 1.6 times fewer than mean-field decoding.
Its accuracy differs from mean-field decoding by $-2.4$ to $+1.8$ points.
On MATH it is 1.3 points lower, and this difference is significant ($p = 0.02$).
Its throughput is 1.10 to 1.32 times that of mean-field decoding, and it is the highest of all methods on GSM8K, MATH, and HumanEval.
On MBPP, DAWN is slightly faster (19.5 against 18.4 tokens per second).

\begin{table}[t]
\centering
\caption{Results on LLaDA-1.5. Our method uses the setting chosen on LLaDA-8B-Instruct, with no new tuning. Columns and TPS measurement as in Table~\ref{tab:main}. Bold marks the best value in each efficiency column.}
\label{tab:llada15}
\small
\setlength{\tabcolsep}{3.5pt}
\begin{tabular}{llcccccc}
\toprule
Bench. & Method & Acc.\ $\uparrow$ & NFE $\downarrow$ & Step red. $\uparrow$ & TPS $\uparrow$ & Time (s) $\downarrow$ & $\Delta$ vs.\ MF [95\% CI] \\
\midrule
\multirow{5}{*}{GSM8K}
 & One token & 80.59 & 256.0 & 1.0$\times$ & 6.9 & 31.82 & \\
 & Fast-dLLM & 81.12 & 75.4 & 3.4$\times$ & 22.5 & 9.70 & \\
 & DAWN & 81.35 & 54.4 & 4.7$\times$ & 30.7 & 7.11 & \\
 & MF & 81.35 & 64.1 & 4.0$\times$ & 28.2 & 7.78 & \\
 & Ours & 80.36 & \textbf{40.2} & \textbf{6.4$\times$} & \textbf{37.3} & \textbf{5.93} & $-1.0$ [$-2.9$, $+0.9$] \\
\midrule
\multirow{5}{*}{MATH}
 & One token & 33.96 & 256.0 & 1.0$\times$ & 8.8 & 25.62 & \\
 & Fast-dLLM & 33.96 & 95.6 & 2.7$\times$ & 22.2 & 10.15 & \\
 & DAWN & 32.20 & 72.7 & 3.5$\times$ & 28.4 & 7.95 & \\
 & MF & 33.28 & 77.3 & 3.3$\times$ & 28.8 & 7.88 & \\
 & Ours & 32.02 & \textbf{49.5} & \textbf{5.2$\times$} & \textbf{37.4} & \textbf{6.09} & $-1.3$ [$-2.3$, $-0.2$] \\
\midrule
\multirow{5}{*}{HumanEval}
 & One token & 43.29 & 256.0 & 1.0$\times$ & 7.1 & 11.77 & \\
 & Fast-dLLM & 43.29 & 99.9 & 2.6$\times$ & 18.3 & 4.56 & \\
 & DAWN & 40.85 & 77.2 & 3.3$\times$ & 23.5 & 3.58 & \\
 & MF & 38.41 & 72.4 & 3.5$\times$ & 25.2 & 3.29 & \\
 & Ours & 40.24 & \textbf{44.4} & \textbf{5.8$\times$} & \textbf{27.8} & \textbf{2.83} & $+1.8$ [$-3.4$, $+7.0$] \\
\midrule
\multirow{5}{*}{MBPP}
 & One token & 37.40 & 256.0 & 1.0$\times$ & 2.4 & 26.58 & \\
 & Fast-dLLM & 37.40 & 42.1 & 6.1$\times$ & 14.1 & 4.49 & \\
 & DAWN & 35.80 & 30.0 & 8.5$\times$ & \textbf{19.5} & \textbf{3.29} & \\
 & MF & 39.80 & 38.3 & 6.7$\times$ & 16.2 & 3.96 & \\
 & Ours & 37.40 & \textbf{28.5} & \textbf{9.0$\times$} & 18.4 & 3.58 & $-2.4$ [$-6.0$, $+1.2$] \\
\bottomrule
\end{tabular}
\end{table}

Table~\ref{tab:dream} reports Dream-v0-Instruct-7B, with the setting calibrated on Dream.
Our method needs 1.7 to 2.2 times fewer steps than mean-field decoding on every benchmark.
It is less accurate than mean-field decoding on all four benchmarks, by 0.6 to 6.1 points.
The largest drop is on HumanEval, where the confidence interval is wide (164 problems), and the drop on MATH (2.1 points) is significant ($p < 0.001$).
It has the shortest time per answer on GSM8K, MATH, and HumanEval, and on MBPP DAWN is slightly faster.
Tokens per second counts only tokens before the end-of-text token, and our answers end earlier (Appendix~\ref{app:latency}), so our gain shows more in time per answer.

\paragraph{Calibration narrows the drops.}\label{sec:dreamcal} We keep one setting per model family, chosen like mean-field decoding on GSM8K training questions, and $\tau$ trades steps for accuracy. Stricter thresholds narrow the drops and keep most of the step savings: on MATH, $\tau = 0.99$ on Dream and $0.95$ on LLaDA-1.5 are within 0.8 and 0.1 points of mean-field decoding (95\% CI $[-1.9, +0.3]$ and $[-1.2, +0.9]$, so at most 2 points lower) with 39\% and 29\% fewer steps (Appendix~\ref{app:math}). On Dream HumanEval, $\tau = 0.97$ ($K = 12$) narrows the gap to 3.7 points (CI $[-10.8, +3.5]$) with 45\% fewer steps (Appendices~\ref{app:curves} and~\ref{app:curves_dream}).

\begin{table}[t]
\centering
\caption{Results on Dream-v0-Instruct-7B. Our method uses the setting calibrated on Dream ($K = 20$, $\tau = 0.95$). Columns and measurement as in Table~\ref{tab:main}. Bold marks the best value in each efficiency column.}
\label{tab:dream}
\small
\setlength{\tabcolsep}{3.5pt}
\begin{tabular}{llcccccc}
\toprule
Bench. & Method & Acc.\ $\uparrow$ & NFE $\downarrow$ & Step red. $\uparrow$ & TPS $\uparrow$ & Time (s) $\downarrow$ & $\Delta$ vs.\ MF [95\% CI] \\
\midrule
\multirow{5}{*}{GSM8K}
 & One token & 75.06 & 256.0 & 1.0$\times$ & 4.8 & 27.17 &  \\
 & Fast-dLLM & 74.98 & 62.2 & 4.1$\times$ & 19.1 & 6.71 &  \\
 & DAWN & 72.25 & 52.0 & 4.9$\times$ & \textbf{22.3} & 5.79 &  \\
 & MF & 76.19 & 69.2 & 3.7$\times$ & 17.0 & 7.41 &  \\
 & Ours & 75.13 & \textbf{41.4} & \textbf{6.2$\times$} & 21.1 & \textbf{5.67} & $-1.1$ [$-3.2$, $+1.1$] \\
\midrule
\multirow{5}{*}{MATH}
 & One token & 38.18 & 256.0 & 1.0$\times$ & 11.2 & 22.37 &  \\
 & Fast-dLLM & 38.34 & 115.0 & 2.2$\times$ & 23.5 & 10.65 &  \\
 & DAWN & 37.70 & 94.4 & 2.7$\times$ & 27.9 & 8.82 &  \\
 & MF & 37.88 & 100.0 & 2.6$\times$ & 21.6 & 9.03 &  \\
 & Ours & 35.76 & \textbf{51.9} & \textbf{4.9$\times$} & \textbf{29.4} & \textbf{5.97} & $-2.1$ [$-3.2$, $-1.0$] \\
\midrule
\multirow{5}{*}{HumanEval}
 & One token & 52.44 & 256.0 & 1.0$\times$ & 16.3 & 10.44 &  \\
 & Fast-dLLM & 55.49 & 90.7 & 2.8$\times$ & 38.8 & 3.83 &  \\
 & DAWN & 54.27 & 75.5 & 3.4$\times$ & \textbf{46.2} & 3.16 &  \\
 & MF & 54.27 & 76.0 & 3.4$\times$ & 35.7 & 3.21 &  \\
 & Ours & 48.17 & \textbf{40.1} & \textbf{6.4$\times$} & 34.0 & \textbf{2.59} & $-6.1$ [$-13.2$, $+1.0$] \\
\midrule
\multirow{5}{*}{MBPP}
 & One token & 54.00 & 256.0 & 1.0$\times$ & 5.5 & 20.39 &  \\
 & Fast-dLLM & 56.00 & 63.6 & 4.0$\times$ & 26.9 & 5.17 &  \\
 & DAWN & 56.80 & 42.9 & 6.0$\times$ & \textbf{30.1} & \textbf{3.55} &  \\
 & MF & 54.20 & 77.2 & 3.3$\times$ & 25.6 & 6.22 &  \\
 & Ours & 53.60 & \textbf{34.4} & \textbf{7.4$\times$} & 18.6 & 3.79 & $-0.6$ [$-4.0$, $+2.8$] \\
\bottomrule
\end{tabular}
\end{table}

\subsection{Where the savings come from}
\label{sec:analysis}

Appendix~\ref{app:pairwise} isolates the pairwise term at a common threshold.
Our coupling needs clearly fewer steps than both the confidence-only rule and mean-field decoding, because it detects support: tokens that confirm each other are committed together.
Table~\ref{tab:controls} tests whether the measured pairs matter, or whether any added support would do.
Each control changes only the coupling of our method, on the same measured pairs, with all other settings fixed.
Shuffling the measured values across pairs keeps their distribution and breaks accuracy, by 36 points on GSM8K and 20 on MBPP.
A constant support bonus, tuned on held-out GSM8K questions, keeps accuracy and needs 44\% to 46\% more steps, and keeping the measured conflicts with a tuned constant support needs 53\% to 72\% more.
The gain therefore comes from knowing which pairs support each other.
The two signs work together: keeping only support collapses accuracy, and keeping only conflicts needs about as many steps as mean-field decoding.
Averaging the two directions where both are measured costs 2 to 3 points (significant on MBPP), so its direction carries some information.

\begin{table}[t]
\centering
\caption{Controls on the measured coupling, LLaDA-8B-Instruct, full test sets (accuracy \% / NFE). Each row changes only the coupling of our method on the measured pairs. $\Delta$ is the accuracy difference from our method with a paired 95\% confidence interval. The constants ($b = 0.02$ and $0.05$) were chosen on 100 GSM8K training questions. Appendix~\ref{app:controls} defines each control.}
\label{tab:controls}
\small
\setlength{\tabcolsep}{3.5pt}
\begin{tabular}{lcccc}
\toprule
 & \multicolumn{2}{c}{GSM8K} & \multicolumn{2}{c}{MBPP} \\
\cmidrule(lr){2-3} \cmidrule(lr){4-5}
Coupling & Acc.\ / NFE & $\Delta$ [95\% CI] & Acc.\ / NFE & $\Delta$ [95\% CI] \\
\midrule
Measured (ours) & 76.27 / 40.1 & & 29.00 / 32.3 & \\
Shuffled across pairs & 40.11 / 44.7 & $-36.2$ [$-39.1$, $-33.2$] & 8.60 / 30.9 & $-20.4$ [$-24.2$, $-16.6$] \\
Support only & 53.37 / 39.8 & $-22.9$ [$-25.7$, $-20.1$] & 15.40 / 29.3 & $-13.6$ [$-16.9$, $-10.3$] \\
Conflict only & \textbf{78.24} / 72.2 & $+2.0$ [$-0.2$, $+4.1$] & 32.20 / 51.9 & $+3.2$ [$+0.0$, $+6.4$] \\
Partially symmetrized & 74.22 / \textbf{37.4} & $-2.1$ [$-4.1$, $+0.0$] & 26.00 / 30.8 & $-3.0$ [$-5.5$, $-0.5$] \\
Constant bonus & 76.72 / 58.6 & $+0.5$ [$-1.7$, $+2.6$] & 29.80 / 46.4 & $+0.8$ [$-2.2$, $+3.8$] \\
Conflicts + constant support & 77.33 / 68.9 & $+1.1$ [$-1.1$, $+3.2$] & \textbf{32.80} / 49.3 & $+3.8$ [$+0.7$, $+6.9$] \\
\bottomrule
\end{tabular}
\end{table}

The coupling matters most when decoding is pushed hard.
At about 34 steps on GSM8K, the confidence-only rule drops to 62\% accuracy, while our method keeps 75\% (Figure~\ref{fig:curves}).

\subsection{Ablations}

Appendix~\ref{app:abl} varies the number of probes and the block size on GSM8K.
More probes save steps at the same accuracy.
Larger blocks need more probes to keep the same share of candidates measured; with that, block 32 matches mean-field accuracy with far fewer steps, while block 64 loses accuracy.
From one to four mean-field iterations, our steps change little and no accuracy difference is significant, while mean-field decoding alternates between short and long schedules with the parity of $R$ (Appendix~\ref{app:iters}).

\subsection{Probe fidelity and cost}
\label{sec:fidelity}

Our probes agree closely with exact full-sequence probes.
During greedy decoding on GSM8K and HumanEval, the two couplings have a Pearson correlation of 0.96 to 0.97, and they agree in sign on 97\% of the pairs with a clear effect ($|\Delta_{ij}| > 0.5$; Appendix~\ref{app:impl}).
About half of the candidate pairs are measured (51\% on GSM8K, 55\% on MBPP; the rest keep $\Delta_{ij} = 0$), and 31\% and 23\% of probes are dropped because their token changed.
On LLaDA, at a similar step budget, a confidence threshold commits three to four times as many tokens that are overturned by the rest of their commit set (Appendix~\ref{app:decisions}).
The probes are also cheap. Exact probes cost about nine forward passes per step, and a separate probe call about half a pass, mostly for reading the weights. Inside the main pass, a step costs 1.2 to 1.5 times a mean-field step, and our method needs 26\% to 55\% fewer steps.
A variant that learns the coupling from the decoder's next step, with no probes, saves fewer steps (Appendix~\ref{app:online}).

\section{Related Work}

Confidence thresholds \citep{wu2025fastdllm}, entropy bounds, and dilated schedules \citep{benhamu2025ebsampler,luxembourg2026dus} decide how many positions to commit at once.
Mean-field decoding adds a similarity penalty and a variational update \citep{zoabi2026meanfield}.
DAWN and dependency-aware parallel decoding build a dependency graph from attention maps and choose an independent set \citep{luo2026dawn,dapd2026}.
DEMASK trains a predictor of the total-variation effect of revealing a token and gives a guarantee on the joint error \citep{demask2026}; our coupling is close to this quantity, and its predictor is not public.
AXON reveals supportive anchors on top of an existing decoder, approximating their value with attention \citep{ayoub2026axon}; it could run on top of ours, and our probes measure a similar quantity directly.
PUNT tests independence with extra model calls and prunes every dependent token \citep{azangulov2025punt}. Conditional influence is therefore not new; we add a training-free way to measure it online, with sign and direction, cheaply enough to coordinate parallel commits.
Other work verifies drafts with an autoregressive model or with the model itself \citep{israel2025apd,han2026s2d2}, verifies drafted future steps \citep{pan2026trajectory,shlomi2026spot}, or remasks committed tokens \citep{wang2025remdm}; these ideas are complementary to ours.
In speculative decoding, DCOCO trains a small module to score adjacent draft tokens \citep{eisenstadt2026dcoco}; we measure such dependencies without training.

\section{Limitations}

A step with probes costs 1.2 to 1.5 times a mean-field step (1.4 to 1.5 times with the cache), so the wall-clock gain is smaller than the step gain; a leaner probe kernel or cheaper coupling estimation would widen it.
The selected settings trade accuracy for speed on MATH and Dream HumanEval (Section~\ref{sec:dreamcal}).
Mutual support shows self-consistency, not correctness, so wrong tokens can reinforce each other. All answers have 256 tokens. With a key-value cache (Appendix~\ref{app:cache}), our GSM8K accuracy is below both baselines, and at block 20 we are slightly slower than mean-field decoding.

\section{Conclusion}

The pairwise term of parallel decoding can be measured instead of approximated, cheaply, by probes inside the main pass. The measured coupling shows support as well as conflict, and on three models it needs far fewer steps than mean-field decoding with no training, with and without a key-value cache. Shuffled couplings and a uniform bonus both lose this gain.

\bibliographystyle{plainnat}
\bibliography{refs}

@article{zoabi2026meanfield,
  title   = {Mean-Field Parallel Decoding for Discrete Diffusion Language Models},
  author  = {Zoabi, Tamim and Ali, Ameen and Ringel, Liran and Wolf, Lior},
  journal = {arXiv preprint arXiv:2606.15805},
  year    = {2026}
}

@article{demask2026,
  title   = {Dependency-Guided Parallel Decoding in Discrete Diffusion Language Models},
  author  = {Ringel, Liran and Ali, Ameen and Romano, Yaniv},
  journal = {arXiv preprint arXiv:2604.02560},
  year    = {2026}
}

@article{luo2026dawn,
  title   = {{DAWN}: Dependency-Aware Fast Inference for Diffusion {LLMs}},
  author  = {Luo, Lizhuo and Shi, Zhuoran and Luo, Jiajun and Wang, Zhi and Ren, Shen and Wang, Wenya and Zhang, Tianwei},
  journal = {arXiv preprint arXiv:2602.06953},
  year    = {2026}
}

@article{dapd2026,
  title   = {{DAPD}: Dependency-Aware Parallel Decoding via Attention for Diffusion {LLMs}},
  author  = {Kim, Bumjun and Jeon, Dongjae and Jeon, Moongyu and No, Albert},
  journal = {arXiv preprint arXiv:2603.12996},
  year    = {2026}
}

@article{nie2025llada,
  title={Large language diffusion models},
  author={Nie, Shen and Zhu, Fengqi and You, Zebin and Zhang, Xiaolu and Ou, Jingyang and Hu, Jun and Zhou, Jun and Lin, Yankai and Wen, Ji-Rong and Li, Chongxuan},
  journal={Advances in Neural Information Processing Systems},
  volume={38},
  pages={50608--50646},
  year={2026}
}

@inproceedings{wu2025fastdllm,
  title={{Fast-dLLM}: Training-free acceleration of diffusion {LLM} by enabling {KV} cache and parallel decoding},
  author={Wu, Chengyue and Zhang, Hao and Xue, Shuchen and Liu, Zhijian and Diao, Shizhe and Zhu, Ligeng and Luo, Ping and Han, Song and Xie, Enze},
  booktitle={International Conference on Learning Representations},
  volume={2026},
  pages={57027--57051},
  year={2026}
}

@article{chen2021codex,
  title   = {Evaluating Large Language Models Trained on Code},
  author  = {Chen, Mark and others},
  journal = {arXiv preprint arXiv:2107.03374},
  year    = {2021}
}

@article{austin2021mbpp,
  title   = {Program Synthesis with Large Language Models},
  author  = {Austin, Jacob and others},
  journal = {arXiv preprint arXiv:2108.07732},
  year    = {2021}
}

@article{cobbe2021gsm8k,
  title   = {Training Verifiers to Solve Math Word Problems},
  author  = {Cobbe, Karl and others},
  journal = {arXiv preprint arXiv:2110.14168},
  year    = {2021}
}

@article{achiam2023gpt4,
  title   = {{GPT}-4 Technical Report},
  author  = {Achiam, Josh and Adler, Steven and Agarwal, Sandhini and others},
  journal = {arXiv preprint arXiv:2303.08774},
  year    = {2023}
}

@article{bai2023qwen,
  title   = {Qwen Technical Report},
  author  = {Bai, Jinze and Bai, Shuai and Chu, Yunfei and others},
  journal = {arXiv preprint arXiv:2309.16609},
  year    = {2023}
}

@article{comanici2025gemini,
  title   = {Gemini 2.5: Pushing the Frontier with Advanced Reasoning, Multimodality, Long Context, and Next Generation Agentic Capabilities},
  author  = {Comanici, Gheorghe and Bieber, Eric and Schaekermann, Mike and others},
  journal = {arXiv preprint arXiv:2507.06261},
  year    = {2025}
}

@article{liu2024deepseek,
  title   = {{DeepSeek-V3} Technical Report},
  author  = {Liu, Aixin and Feng, Bei and Xue, Bing and others},
  journal = {arXiv preprint arXiv:2412.19437},
  year    = {2024}
}

@inproceedings{austin2021d3pm,
  title     = {Structured Denoising Diffusion Models in Discrete State-Spaces},
  author    = {Austin, Jacob and Johnson, Daniel D. and Ho, Jonathan and Tarlow, Daniel and van den Berg, Rianne},
  booktitle = {Advances in Neural Information Processing Systems},
  year      = {2021}
}

@inproceedings{li2022diffusionlm,
  title     = {Diffusion-{LM} Improves Controllable Text Generation},
  author    = {Li, Xiang Lisa and Thickstun, John and Gulrajani, Ishaan and Liang, Percy and Hashimoto, Tatsunori},
  booktitle = {Advances in Neural Information Processing Systems},
  year      = {2022}
}

@inproceedings{zhu2025llada15,
  title={{LLaDA} 1.5: Variance-reduced preference optimization for large language diffusion models},
  author={Zhu, Fengqi and Wang, Rongzhen and Nie, Shen and Zhang, Xiaolu and Wu, Chunwei and Zhou, Jun and Lin, Yankai and Wen, Ji-Rong and Li, Chongxuan},
  booktitle={Proceedings of the 64th Annual Meeting of the Association for Computational Linguistics (Volume 1: Long Papers)},
  pages={11425--11460},
  year={2026}
}

@article{ye2025dream,
  title   = {Dream 7{B}: Diffusion Large Language Models},
  author  = {Ye, Jiacheng and Xie, Zhihui and Zheng, Lin and Gao, Jiahui and Wu, Zirui and Jiang, Xin and Li, Zhenguo and Kong, Lingpeng},
  journal = {arXiv preprint arXiv:2508.15487},
  year    = {2025}
}

@article{bie2025llada20,
  title   = {{LLaDA2.0}: Scaling Up Diffusion Language Models to 100{B}},
  author  = {Bie, Tiwei and Cao, Maosong and Chen, Kun and others},
  journal = {arXiv preprint},
  year    = {2025}
}

@article{bie2026llada21,
  title   = {{LLaDA2.1}: Speeding Up Text Diffusion via Token Editing},
  author  = {Bie, Tiwei and Cao, Maosong and Cao, Xiang and others},
  journal = {arXiv preprint},
  year    = {2026}
}

@article{zhang2026order,
  title   = {Generation Order and Parallel Decoding in Masked Diffusion Models: An Information-Theoretic Perspective},
  author  = {Zhang, Shaorong and Yu, Longxuan and Brekelmans, Rob and Tang, Luhan and Asif, Salman and Ver Steeg, Greg},
  journal = {arXiv preprint arXiv:2602.00286},
  year    = {2026}
}

@inproceedings{hendrycks2021math,
  title     = {Measuring Mathematical Problem Solving With the {MATH} Dataset},
  author    = {Hendrycks, Dan and Burns, Collin and Kadavath, Saurav and Arora, Akul and Basart, Steven and Tang, Eric and Song, Dawn and Steinhardt, Jacob},
  booktitle = {Advances in Neural Information Processing Systems, Datasets and Benchmarks Track},
  year      = {2021}
}

@inproceedings{lewkowycz2022minerva,
  title     = {Solving Quantitative Reasoning Problems with Language Models},
  author    = {Lewkowycz, Aitor and Andreassen, Anders and Dohan, David and Dyer, Ethan and Michalewski, Henryk and Ramasesh, Vinay and Slone, Ambrose and Anil, Cem and Schlag, Imanol and Gutman-Solo, Theo and Wu, Yuhuai and Neyshabur, Behnam and Gur-Ari, Guy and Misra, Vedant},
  booktitle = {Advances in Neural Information Processing Systems},
  year      = {2022}
}

@inproceedings{benhamu2025ebsampler,
  title     = {Accelerated Sampling from Masked Diffusion Models via Entropy Bounded Unmasking},
  author    = {Ben-Hamu, Heli and Gat, Itai and Severo, Daniel and Nolte, Niklas and Karrer, Brian},
  booktitle = {Advances in Neural Information Processing Systems},
  year      = {2025}
}

@inproceedings{luxembourg2026dus,
  title     = {Plan for Speed: Dilated Scheduling for Masked Diffusion Language Models},
  author    = {Luxembourg, Omer and Permuter, Haim and Nachmani, Eliya},
  booktitle = {International Conference on Machine Learning},
  year      = {2026}
}

@article{israel2025apd,
  title={Accelerating diffusion {LLMs} via adaptive parallel decoding},
  author={Israel, Daniel and Van den Broeck, Guy and Grover, Aditya},
  journal={Advances in neural information processing systems},
  volume={38},
  pages={52870--52888},
  year={2026}
}

@article{han2026s2d2,
  title   = {{S2D2}: Fast Decoding for Diffusion {LLMs} via Training-Free Self-Speculation},
  author  = {Han, Ligong and Wang, Hao and Gao, Han and Xu, Kai and Srivastava, Akash},
  journal = {arXiv preprint arXiv:2603.25702},
  year    = {2026}
}

@article{wang2025remdm,
  title={Remasking discrete diffusion models with inference-time scaling},
  author={Wang, Guanghan and Schiff, Yair and Sahoo, Subham and Kuleshov, Volodymyr},
  journal={Advances in Neural Information Processing Systems},
  volume={38},
  pages={147282--147339},
  year={2026}
}

@article{eisenstadt2026dcoco,
  title   = {Contextual Correction of Parallel Drafts for Speculative Decoding},
  author  = {Eisenstadt, Roy and Lidar, Harel and Gonen, Tal and Luxembourg, Omer and Zimerman, Itamar and Wolf, Lior and David, Eli and Naiman, Ilan},
  journal = {arXiv preprint arXiv:2610.12570},
  year    = {2026}
}

@article{ayoub2026axon,
  title   = {Supportive Token Revealing for Fast Diffusion Language Model Decoding},
  author  = {Abu Ayoub, Giries and Barbara, Mario and Pastor-P{\'e}rez, Llu{\'i}s and Bien, Tanja and Barthakur, Aneesh and Maalouf, Alaa and Mualem, Loay},
  journal = {arXiv preprint arXiv:2606.04236},
  year    = {2026}
}

@article{pan2026trajectory,
  title={Trajectory-Level Speculative Decoding for Diffusion Language Models},
  author={Pan, Tianxiang and Gong, Baitao and Guang, Mo and Yong, Hongwei and Jiang, Tianpeng and Li, Yaqian and Cao, Zheng and Long, Kaiwen},
  journal={arXiv preprint arXiv:2608.27514},
  year={2026}
}

@article{shlomi2026spot,
  title={On the {SPOT}: Enhancing Speculative Decoding for Diffusion Language Models via Commit-Order Prediction},
  author={Shlomi, Eliezer and Rahimi, Eliron and Mendelson, Avi and Baskin, Chaim},
  journal={arXiv preprint arXiv:2610.22543},
  year={2026}
}

@inproceedings{azangulov2025punt,
  title={Parallel sampling from masked diffusion models via conditional independence testing},
  author={Azangulov, Iskander and Pandeva, Teodora and Prasad, Niranjani and Zazo, Javier and Karmalkar, Sushrut},
  booktitle={International Conference on Learning Representations},
  volume={2026},
  pages={140485--140521},
  year={2026}
}

\appendix
\section{Implementation details}
\label{app:impl}
A probe row holds one copy of the still-masked positions of the current block, with the probed candidate $j$ set to its proposal $v_j$ and the other positions left masked.
In every layer, the main tokens and all probe rows go through the same normalization, projections, MLP, and output head in one batch.
Each probe query attends to the main keys and values of the same layer at every position it does not recompute, and to the keys and values of its own row at the positions it does; it never sees another probe row.
The main keys at the recomputed positions are masked out for the probe queries.
Rotary embeddings use each position's index in the sequence, so a probe token sits where it would in the main pass.
Main tokens use the model's usual bidirectional attention, so the main logits match a plain forward pass up to bfloat16 rounding.

The fidelity numbers in Section~\ref{sec:fidelity} come from 3 test questions of GSM8K (5-shot) and 3 of HumanEval, decoded greedily for three blocks of 20; we measure at the start and middle of each block, giving 18 states and 4{,}230 pairs per task.
The contraction statistics in Section~\ref{sec:contraction} come from 1{,}805 steps on GSM8K and 1{,}662 on HumanEval.
With $K = 12$, a fused step uses 0.26\,GB more peak memory than a plain step on GSM8K and 0.14\,GB more on HumanEval, on top of 14.9\,GB of weights.
Timings come from one job per benchmark on one A100 (80\,GB), with batch size 1 and bfloat16. Where our method is timed at the start and end of a job, the two times differ by at most 1.3\%, except in one LLaDA-1.5 GSM8K job (16\%); for that benchmark we report the median over three jobs.

\section{Effect of the pairwise term}
\label{app:pairwise}

Table~\ref{tab:ablation} isolates the pairwise term.
All rows use the same block size, schedule, and threshold $\tau = 0.85$, the published threshold of mean-field decoding, so the pairwise term is the only difference.
The first row has no pairwise term, so each position is committed on its own confidence.
The second row is mean-field decoding with the similarity coupling, as in Table~\ref{tab:main}.
The third row is our method with $\tau = 0.85$ in place of its selected $\tau = 0.9$, which is why its numbers differ from Table~\ref{tab:main}.

\begin{table}[h]
\centering
\caption{Effect of the pairwise term (accuracy \% / NFE). All rows use block 20 and $\tau = 0.85$, so only the pairwise term changes. Our row uses the selected probes ($K = 12$, inside the main pass) at this common threshold.}
\label{tab:ablation}
\begin{tabular}{lccc}
\toprule
Pairwise term & GSM8K & HumanEval & MBPP \\
\midrule
None (confidence only) & 78.09 / 52.1 & 35.98 / 49.6 & 26.80 / 42.7 \\
Similarity $D$ (MF) & 76.57 / 65.4 & 37.20 / 61.7 & 30.20 / 51.6 \\
Measured $\Delta$ (ours) & 76.42 / \textbf{37.4} & 34.76 / \textbf{37.7} & 28.60 / \textbf{30.7} \\
\bottomrule
\end{tabular}
\end{table}

The similarity term makes decoding more careful: it adds steps relative to the confidence-only rule, and it helps accuracy on MBPP ($+3.4$ points, $p = 0.027$).
The measured coupling moves in the other direction on steps.
At the same threshold it needs 24\% to 43\% fewer steps than either rule.

\section{Accuracy against time per answer}
\label{app:latency}
\begin{figure}[h]
\centering
\includegraphics[width=\linewidth]{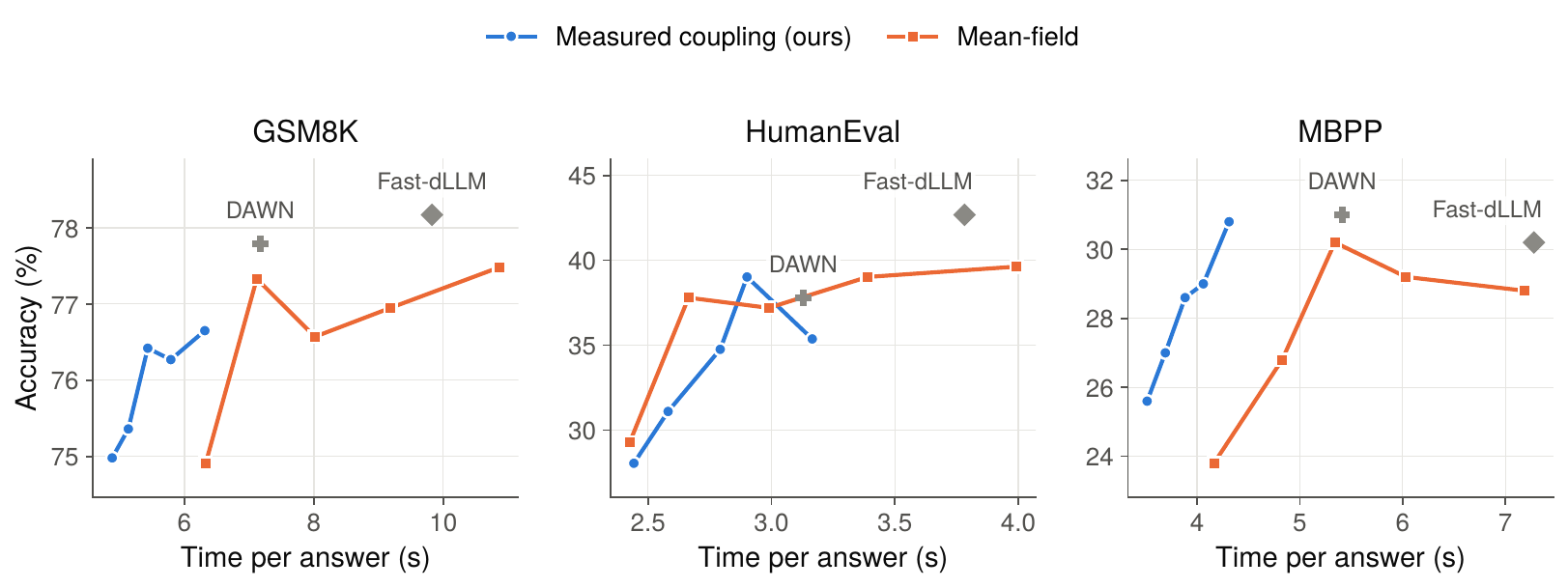}
\caption{Accuracy against wall-clock time per answer on LLaDA-8B-Instruct. Curves sweep $\tau$ from 0.75 to 0.95 for our method and for mean-field decoding. Gray points are Fast-dLLM and DAWN at their published settings. All points of a panel are timed on the same GPU.}
\label{fig:latency}
\end{figure}

Figure~\ref{fig:latency} sweeps the threshold of our method and of mean-field decoding on LLaDA-8B-Instruct and plots accuracy against wall-clock time per answer.
All points of a benchmark are timed in one job on the same A100 GPU (GSM8K on its first 200 questions, HumanEval on all 164, MBPP on its first 200), with batch size 1 and bfloat16; accuracy is from the full test sets.
We timed our method at $\tau = 0.9$ at the start and at the end of each job, and the two times differ by 0.6\% to 3.1\%.
On GSM8K and MBPP our curve lies to the left of the mean-field curve; on GSM8K it stops at 76.7\%, slightly below the best mean-field point (77.5\%).
On HumanEval the curves cross: our method is faster at the higher accuracies and mean-field decoding at the lower ones.
Our answers are also shorter. Counting tokens before the end-of-text token, they are 5\% to 56\% shorter than those of mean-field decoding on Dream, 17\% shorter on LLaDA-8B MBPP, and within about 5\% in the other LLaDA settings.

\section{Results with a key-value cache}
\label{app:cache}
We also run every method with the dual cache of Fast-dLLM \citep{wu2025fastdllm}. A full forward pass at the start of each block fills the cache, and every later step of the block runs only the block against it. All three decoders share one implementation of this step, captured as a CUDA graph, with one fused attention kernel that applies the rotary embedding on load; our probe rows run in the same pass and kernel. All methods use block 32, where the block-start passes cost the same for every method.
Table~\ref{tab:cache} reports our method with $K = 4$ probes and $\tau = 0.9$ on LLaDA-8B-Instruct. Accuracy is from the full test sets, and time is measured for all methods in one job on the same GPU after a warm-up of 20 questions (GSM8K on its first 200 questions, MBPP on all 500).
Our method needs 29\% to 31\% fewer steps than mean-field decoding and 49\% to 55\% fewer than Fast-dLLM, and has the shortest time per answer on both benchmarks. Its accuracy is higher than both baselines on MBPP, by 4.0 points over mean-field decoding ($p = 0.03$), and lower on GSM8K.
This setting is the fastest of the cached grid we ran ($K \in \{4, 8\}$, $\tau \in \{0.85, 0.9\}$, blocks 20 and 32). The setting chosen by our rule on GSM8K training questions uses block 20 and $K = 8$ (GSM8K 75.1\%, MBPP 27.6\%); at block 20 every method pays more block starts, and ours is 3\% to 5\% slower than mean-field decoding at the same block.

\begin{table}[h]
\centering
\caption{Methods with the dual key-value cache on LLaDA-8B-Instruct, block 32 (full test sets). Time is the wall-clock time per answer in seconds. Step time excludes the full passes at block starts, which cost about the same for every method (1.04\,s per answer on GSM8K, 0.86 to 0.88\,s on MBPP).}
\label{tab:cache}
\small
\setlength{\tabcolsep}{3.5pt}
\begin{tabular}{llcccc}
\toprule
Bench. & Method & Acc.\ $\uparrow$ & NFE $\downarrow$ & Time (s) $\downarrow$ & Step time (s) $\downarrow$ \\
\midrule
\multirow{3}{*}{GSM8K}
 & Fast-dLLM & 79.38 & 85.6 & 2.31 & 1.27 \\
 & MF & 77.03 & 61.6 & 2.06 & 1.02 \\
 & Ours & 75.44 & \textbf{43.6} & \textbf{1.96} & \textbf{0.92} \\
\midrule
\multirow{3}{*}{MBPP}
 & Fast-dLLM & 26.60 & 68.5 & 1.84 & 0.96 \\
 & MF & 24.00 & 44.9 & 1.56 & 0.70 \\
 & Ours & \textbf{28.00} & \textbf{30.9} & \textbf{1.51} & \textbf{0.65} \\
\bottomrule
\end{tabular}
\end{table}

\newpage

\section{Accuracy against steps on LLaDA-8B}
\label{app:curves}
\begin{figure}[h]
\centering
\includegraphics[width=\linewidth]{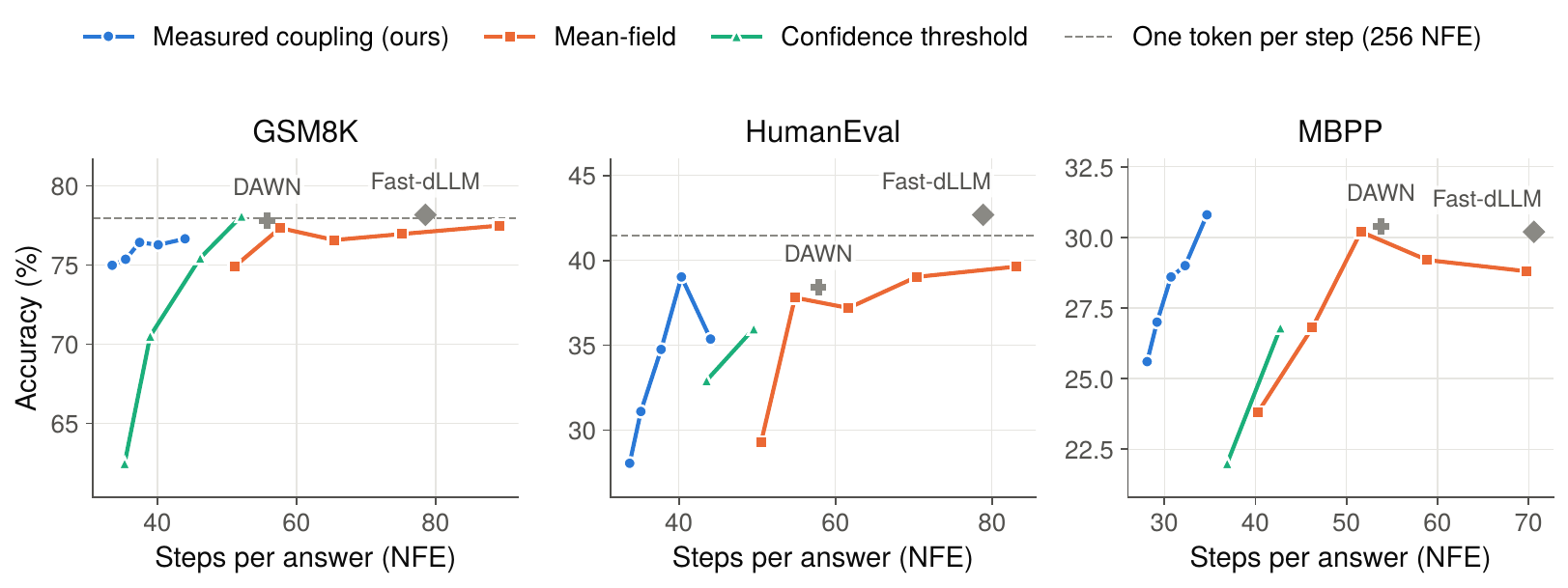}
\caption{Accuracy against the number of steps per answer on LLaDA-8B-Instruct. Curves sweep the threshold $\tau$: 0.75 to 0.95 for our method and for mean-field decoding, and 0.7 to 0.85 for the confidence-only rule (0.8 and 0.85 on code). Gray points are Fast-dLLM and DAWN at their published settings; the dashed line is one-token-per-step decoding.}
\label{fig:curves}
\end{figure}

\section{Accuracy against steps on Dream}
\label{app:curves_dream}
\begin{figure}[h]
\centering
\includegraphics[width=\linewidth]{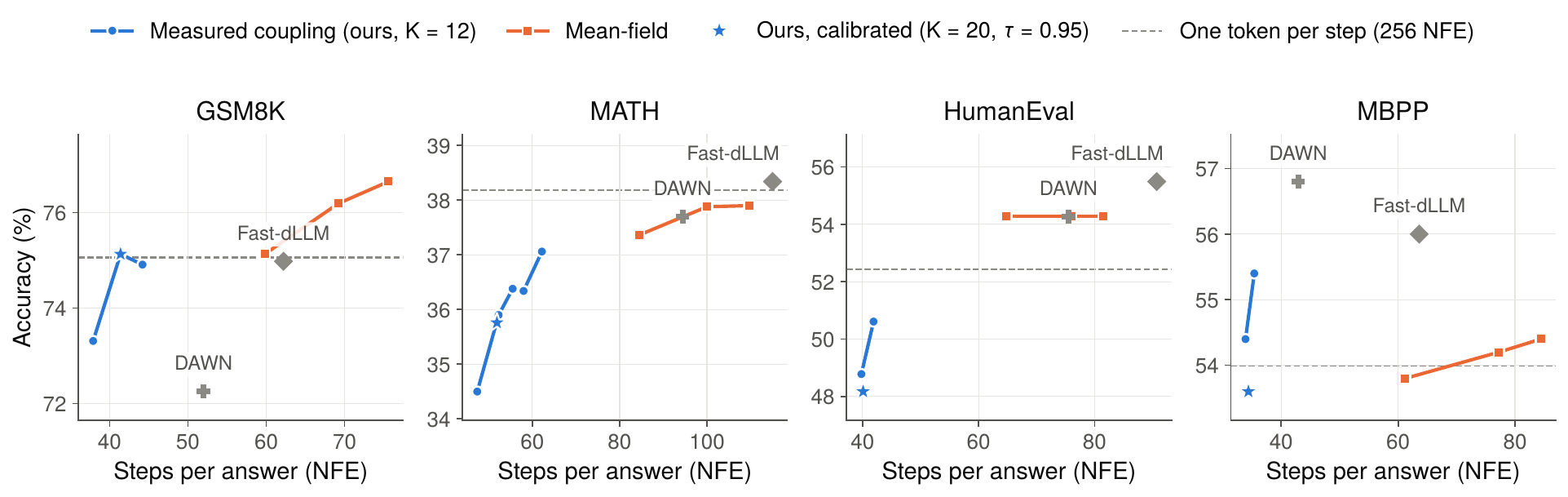}
\caption{Accuracy against steps per answer on Dream-v0-Instruct-7B. Curves sweep $\tau$ over 0.9, 0.95, and 0.97 at block 20, and up to 0.99 for our method on MATH. Our curve uses $K = 12$. The star is our calibrated setting from Table~\ref{tab:dream}. Gray points are Fast-dLLM and DAWN; the dashed line is one-token-per-step decoding.}
\label{fig:curves_dream}
\end{figure}

Figure~\ref{fig:curves_dream} sweeps the threshold on Dream-v0-Instruct-7B. Our curve reaches its accuracy with about half the steps of mean-field decoding, and on MBPP it rises above the mean-field curve.

\section{Where the MATH gap comes from}
\label{app:math}
We paired every MATH problem between our method and mean-field decoding, using the saved answers of the main runs.
The gap is spread over all difficulty levels and subjects instead of concentrating on the hardest problems, where long and sensitive derivations would stress the additive model most.
Repetition loops occur at similar rates for both methods and explain at most 0.2 points of the gap.
On Dream, a stricter threshold narrows the gap: at $\tau = 0.99$ our method comes within 0.8 points of mean-field decoding (37.1\% against 37.9\%, with $K = 12$ and with $K = 20$) with nearly 40\% fewer steps, and the difference is not significant (Figure~\ref{fig:curves_dream}).
The gap therefore depends on the speed setting chosen on held-out data, and a stricter threshold reduces it at the cost of more steps.

\section{Number of probes and block size}
\label{app:abl}
Table~\ref{tab:abl} varies the number of probes $K$ and the block size on GSM8K, with all other settings fixed at the selected ones.

\paragraph{Number of probes.}
Accuracy stays the same for every $K$ we tried, while each extra probe saves steps.
More measured couplings let the decoder commit more tokens together, at no cost in accuracy.

\paragraph{Block size.}
At blocks 16 and 20, our method matches mean-field decoding with far fewer steps, and at block 16 it is slightly more accurate.
With a fixed $K$, larger blocks leave most candidates unprobed: their columns of $\Delta$ are zero, so their influence on other positions is not measured, although they still receive measured influences from the probed candidates. Accuracy drops.
Scaling $K$ with the block size keeps the share of probed candidates constant.
This restores accuracy at block 32, where our method again matches mean-field decoding with far fewer steps.
Very large blocks such as 64 need more than additional probes, and we leave them to future work.

\begin{table}[h]
\centering
\caption{Ablations on GSM8K (accuracy \% / NFE). Left: number of probes $K$ ($\tau = 0.9$, block 20). Right: block size, for our method with $K = 12$, with $K$ scaled with the block ($K = 12$ at blocks 16 and 20, $K = 20$ at 32, $K = 40$ at 64), and for mean-field decoding ($\tau = 0.85$).}
\label{tab:abl}
\small
\begin{tabular}{lc}
\toprule
$K$ & Ours \\
\midrule
4 & 76.42 / 45.0 \\
8 & 76.27 / 41.3 \\
12 & 76.27 / 40.1 \\
20 & 76.27 / \textbf{38.8} \\
\bottomrule
\end{tabular}
\hspace{1.5em}
\begin{tabular}{lccc}
\toprule
Block & Ours ($K = 12$) & Ours (scaled $K$) & MF \\
\midrule
16 & \textbf{77.03} / \textbf{43.2} & 77.03 / 43.2 & 76.95 / 69.0 \\
20 & 76.27 / \textbf{40.1} & 76.27 / 40.1 & 76.57 / 65.4 \\
32 & 72.55 / 34.2 & 76.19 / \textbf{33.6} & 76.42 / 56.4 \\
64 & 67.70 / 29.1 & 67.70 / \textbf{27.4} & 75.44 / 48.3 \\
\bottomrule
\end{tabular}
\end{table}

\section{Definition of the controls}
\label{app:controls}
Every control in Table~\ref{tab:controls} changes only the measured off-diagonal entries of $\Delta$, so it has the same coverage as our method.
Shuffled permutes these values at random with a fixed seed.
Support only sets $\Delta_{ij} \gets \min(\Delta_{ij}, 0)$, and conflict only sets $\Delta_{ij} \gets \max(\Delta_{ij}, 0)$.
Partially symmetrized replaces $\Delta_{ij}$ by $(\Delta_{ij} + \Delta_{ji})/2$ where both directions are measured and keeps the other entries, so the matrix is not fully symmetric.
Constant bonus sets every measured entry to $-b$; conflicts + constant support keeps $\Delta_{ij} > 0$ and sets every measured $\Delta_{ij} < 0$ to $-b$.

\section{Testing the additive model}
\label{app:additivity}
The decoder sums single-token effects over the committed set $S$ (Equation~\ref{eq:additive}).
We test this on the sets it actually commits.
At every step with $|S| \ge 2$, we measure each $\Delta_{ij}$ exactly with one full forward pass per $j \in S$, and the true margin of each remaining position with all of $S$ revealed in one more pass.
A flip means that the additive prediction and the true margin, after the sigmoid, fall on different sides of $\tau$.
Flips are rare for small sets and grow with $|S|$ (Table~\ref{tab:additivity}).
The sum overstates both signs.
Where support dominates, the predicted margin is about 5.5 and the true one about 3.5, so support from several tokens is partly redundant.
Where conflict dominates, the prediction is about $-16$ and the true margin about $-3.5$, so conflicts also saturate.
For the committed tokens themselves, with the rest of their set revealed (Appendix~\ref{app:decisions}), the additive and the exact margin fall on the same side of $\tau$ for 96\% of tokens on GSM8K and 97\% on MBPP.

\begin{table}[h]
\centering
\caption{Additive prediction against the jointly measured margin, LLaDA-8B-Instruct with our method ($K = 12$, $\tau = 0.9$; $K$ scaled with the block at 32 and 64). Share of remaining (uncommitted) positions whose decision at $\tau$ flips, by the size of the committed set. For the committed tokens, see the text.}
\label{tab:additivity}
\small
\begin{tabular}{lccccc}
\toprule
Run & Positions & $|S| = 2$--$3$ & $|S| = 4$--$7$ & $|S| \ge 8$ & All \\
\midrule
GSM8K, block 20 & 7{,}880 & 3.7\% & 7.8\% & 15.6\% & 10.0\% \\
MBPP, block 20 & 7{,}981 & 4.5\% & 7.3\% & 14.2\% & 9.4\% \\
GSM8K, block 32 & 8{,}199 & 2.4\% & 5.7\% & 9.2\% & 7.0\% \\
GSM8K, block 64 & 13{,}147 & 1.4\% & 3.7\% & 10.8\% & 7.2\% \\
\bottomrule
\end{tabular}
\end{table}

\section{Commit decisions}
\label{app:decisions}
We decode 50 test questions of GSM8K and of MBPP with our method and, at every step, compute the commit set on the same state with other couplings (Table~\ref{tab:decisions}).
Block-local probes for all candidates change 4\% of the decisions of exact probes, and measuring only the probed columns accounts for most of the rest.
Most changed decisions hold back a token that exact probes would commit, which costs steps more than accuracy.

We then check each committed token against the rest of its commit set: we reveal the other tokens of the set, leave the token masked, and test whether its committed value is still the most likely one (Table~\ref{tab:overturn}).
End-of-text padding is excluded.
At the same number of steps, a confidence threshold commits three to four times as many overturned tokens as our method, and the gap is largest for small sets.
Slower decoders with 50\% to 65\% more steps are overturned slightly less often.
On Dream ($K = 20$, $\tau = 0.95$), the threshold at a similar step budget is overturned 3.6 times as often on MATH and 1.4 times as often on HumanEval.

\begin{table}[h]
\centering
\caption{Commit decisions on the same decoding states, against exact full-sequence probes for every candidate (50 test questions each, our method with $K = 12$ and $\tau = 0.9$; steps with only end-of-text padding left are excluded).}
\label{tab:decisions}
\small
\begin{tabular}{lcccc}
\toprule
 & \multicolumn{2}{c}{GSM8K} & \multicolumn{2}{c}{MBPP} \\
\cmidrule(lr){2-3} \cmidrule(lr){4-5}
Coupling & Same set & Decisions changed & Same set & Decisions changed \\
\midrule
Block-local, all candidates & 70\% & 4.2\% & 71\% & 4.4\% \\
Exact, probed columns only & 55\% & 10.6\% & 55\% & 11.2\% \\
Ours (inside the main pass) & 50\% & 11.8\% & 47\% & 12.5\% \\
No coupling & 33\% & 27.3\% & 36\% & 25.2\% \\
\bottomrule
\end{tabular}
\end{table}

\begin{table}[h]
\centering
\caption{Committed tokens overturned by the rest of their commit set, by set size (50 test questions each; LLaDA-8B-Instruct unless marked Dream). Conf.\ is the confidence threshold.}
\label{tab:overturn}
\small
\begin{tabular}{llcccccc}
\toprule
Task & Decoder & $\tau$ & NFE & $|S| = 2$--$3$ & $|S| = 4$--$7$ & $|S| \ge 8$ & All \\
\midrule
\multirow{4}{*}{GSM8K} & Ours & 0.9 & 41.1 & 0.4\% & 0.8\% & 0.8\% & 0.8\% \\
 & Conf. & 0.75 & 41.9 & 6.6\% & 4.9\% & 1.7\% & 2.9\% \\
 & Conf. & 0.9 & 66.3 & 0.2\% & 0.3\% & 0.2\% & 0.2\% \\
 & MF & 0.85 & 68.0 & 0.5\% & 0.3\% & 0.1\% & 0.2\% \\
\midrule
\multirow{4}{*}{MBPP} & Ours & 0.9 & 35.0 & 1.2\% & 0.8\% & 0.7\% & 0.8\% \\
 & Conf. & 0.8 & 41.5 & 5.0\% & 3.3\% & 1.0\% & 2.3\% \\
 & Conf. & 0.9 & 53.5 & 0.8\% & 0.4\% & 0.1\% & 0.3\% \\
 & MF & 0.85 & 55.2 & 1.3\% & 0.1\% & 0.0\% & 0.3\% \\
\midrule
\multirow{3}{*}{\shortstack[l]{Dream\\MATH}} & Ours & 0.95 & 60.9 & 0.5\% & 0.5\% & 0.4\% & 0.4\% \\
 & Conf. & 0.8 & 61.6 & 3.4\% & 1.5\% & 1.0\% & 1.5\% \\
 & MF & 0.95 & 116.8 & 0.0\% & 0.0\% & 0.0\% & 0.0\% \\
\midrule
\multirow{3}{*}{\shortstack[l]{Dream\\HumanEval}} & Ours & 0.95 & 28.9 & 0.6\% & 1.1\% & 0.5\% & 0.7\% \\
 & Conf. & 0.85 & 30.3 & 2.7\% & 1.0\% & 0.4\% & 1.0\% \\
 & MF & 0.95 & 56.4 & 0.0\% & 0.0\% & 0.0\% & 0.0\% \\
\bottomrule
\end{tabular}
\end{table}

\section{Number of mean-field iterations}
\label{app:iters}
Table~\ref{tab:iters} varies the number of mean-field iterations $R$ on Dream-v0-Instruct-7B, with all other settings as in Table~\ref{tab:dream}.
For our method, the number of steps stays within 40 to 43 on GSM8K and 34 to 36 on MBPP for every $R$. Accuracy varies by up to 1.2 points on GSM8K and 2.4 points on MBPP, and no difference from $R = 2$ is significant.
Mean-field decoding needs 1.4 to 1.6 times more steps at odd $R$ than at even $R$, which matches the oscillation reported by \citet{zoabi2026meanfield}.
All main results use $R = 2$, as in mean-field decoding.

\begin{table}[h]
\centering
\caption{Number of mean-field iterations $R$ on Dream-v0-Instruct-7B (accuracy \% / NFE). The main results use $R = 2$ (bold).}
\label{tab:iters}
\small
\begin{tabular}{ccccc}
\toprule
 & \multicolumn{2}{c}{Ours ($K = 20$, $\tau = 0.95$)} & \multicolumn{2}{c}{Mean-field ($\tau = 0.95$)} \\
\cmidrule(lr){2-3} \cmidrule(lr){4-5}
& GSM8K & MBPP & GSM8K & MBPP \\
$R$ & Acc.\ \% / NFE & Acc.\ \% / NFE & Acc.\ \% / NFE & Acc.\ \% / NFE \\
\midrule
1 & 75.36 / 42.9 & 56.00 / 36.1 & 75.66 / 113.3 & 57.80 / 118.3 \\
\textbf{2} & \textbf{75.13 / 41.4} & \textbf{53.60 / 34.4} & \textbf{76.19 / 69.2} & \textbf{54.20 / 77.2} \\
3 & 74.15 / 40.2 & 56.00 / 34.3 & 75.89 / 110.3 & 58.40 / 113.9 \\
4 & 74.22 / 39.8 & 54.80 / 33.8 & 75.82 / 70.2 & 54.00 / 79.2 \\
\bottomrule
\end{tabular}
\end{table}

\section{Learned coupling from the next step}
\label{app:online}
\label{sec:online}
The decoder's own next step contains information about the coupling.
After a set $S$ is committed, the next forward pass shows whether each remaining position changes its choice.
This gives an exact label for the event that committing that position together with $S$ would have changed the answer, at no extra cost.
We fit a logistic model of this event on simple features of position pairs (distance and side, similarity of the two predictions, whether the tokens match, and the uncertainty at $i$), by ridge-regularized Newton steps, online and without training data, and commit positions whose predicted error probability is below a risk level $\rho$.

This variant needs no extra forward passes and is useful on its own.
With $\rho$ tuned on GSM8K training questions, it uses fewer steps than mean-field decoding on GSM8K, HumanEval, and MBPP, with accuracy within 1.4 points.
Its predicted flip rates match the observed rates within a few points for every committed-set size. This calibration is about flips, and Appendix~\ref{app:additivity} tests the additive margins directly.
Its labels come only from positions that were not committed, so they rarely cover confident tokens.
Probes reach those tokens directly, which explains the larger savings of measured coupling.

Table~\ref{tab:online} reports this variant on LLaDA-8B-Instruct.
We chose $\rho = 0.05$ on the first 100 GSM8K training questions, the same held-out set as for our main method.
It uses fewer steps than mean-field decoding on all three benchmarks. Its accuracy is 0.2 points higher on GSM8K, equal on HumanEval, and 1.4 points lower on MBPP.

\begin{table}[h]
\centering
\caption{Learned coupling from the next step on LLaDA-8B-Instruct (block 20, $\rho = 0.05$). Mean-field decoding is as in Table~\ref{tab:main}.}
\label{tab:online}
\small
\begin{tabular}{lcccccc}
\toprule
 & \multicolumn{2}{c}{GSM8K} & \multicolumn{2}{c}{HumanEval} & \multicolumn{2}{c}{MBPP} \\
\cmidrule(lr){2-3} \cmidrule(lr){4-5} \cmidrule(lr){6-7}
Method & Acc.\ \% $\uparrow$ & NFE $\downarrow$ & Acc.\ \% $\uparrow$ & NFE $\downarrow$ & Acc.\ \% $\uparrow$ & NFE $\downarrow$ \\
\midrule
Mean-field (MF) & 76.57 & 65.4 & 37.20 & 61.7 & 30.20 & 51.6 \\
Learned coupling & 76.72 & 53.1 & 37.20 & 54.2 & 28.80 & 48.7 \\
\bottomrule
\end{tabular}
\end{table}

\section{Held-out selection results}

\begin{table}[h]
\centering
\caption{Every candidate setting of our method scored on each held-out set. ``Masked only'' probes recompute only the still-masked positions, and ``separate call'' probes run after the main pass instead of inside it. The selected setting, chosen on GSM8K, is in bold. MBPP validation is a check on code. Mean-field decoding is shown for reference only; its setting is the published one.}
\label{tab:heldout}
\small
\begin{tabular}{llccc}
\toprule
Held-out set & Probes & $\tau$ & Accuracy $\uparrow$ & NFE $\downarrow$ \\
\midrule
\multirow{13}{*}{GSM8K train (100)}
 & \textbf{12, inside main pass, masked only} & \textbf{0.90} & \textbf{91.0} & \textbf{39.3} \\
 & 12, inside main pass, masked only & 0.85 & 84.0 & 37.2 \\
 & 8, inside main pass, masked only & 0.90 & 88.0 & 40.1 \\
 & 8, inside main pass, masked only & 0.85 & 86.0 & 38.4 \\
 & 8, separate call & 0.85 & 90.0 & 34.3 \\
 & 8, separate call & 0.90 & 85.0 & 36.2 \\
 & 12, separate call & 0.85 & 84.0 & 32.8 \\
 & 12, separate call & 0.90 & 86.0 & 34.7 \\
 & 20, inside main pass & 0.85 & 84.0 & 35.6 \\
 & 20, inside main pass & 0.90 & 82.0 & 38.0 \\
 & 12, inside main pass, masked only & 0.93 & 86.0 & 42.0 \\
 & 12, inside main pass, masked only & 0.95 & 83.0 & 42.8 \\
 & 12, inside main pass, masked only & 0.97 & 81.0 & 45.5 \\
\midrule
\multirow{13}{*}{MBPP validation (90)}
 & \textbf{12, inside main pass, masked only} & \textbf{0.90} & \textbf{35.56} & \textbf{32.7} \\
 & 8, inside main pass, masked only & 0.90 & 30.00 & 33.5 \\
 & 20, inside main pass, masked only & 0.90 & 30.00 & 33.5 \\
 & 12, inside main pass & 0.90 & 28.89 & 32.9 \\
 & 12, inside main pass & 0.85 & 27.78 & 32.0 \\
 & 20, inside main pass & 0.85 & 27.78 & 31.3 \\
 & 20, inside main pass & 0.90 & 27.78 & 33.5 \\
 & 12, separate call & 0.90 & 26.67 & 29.6 \\
 & 8, separate call & 0.90 & 25.56 & 30.8 \\
 & 12, inside main pass, masked only & 0.93 & 34.44 & 33.8 \\
 & 12, inside main pass, masked only & 0.95 & 37.78 & 35.2 \\
 & 12, inside main pass, masked only & 0.97 & 37.78 & 37.1 \\
 & Mean-field (reference) & 0.85 & 30.00 & 52.2 \\
\bottomrule
\end{tabular}
\end{table}

\begin{table}[h]
\centering
\caption{Held-out candidates on Dream-v0-Instruct-7B, with the same held-out sets and rule as Table~\ref{tab:heldout}. The selected setting, chosen on GSM8K, is in bold; MBPP validation selects the same one. Mean-field decoding is shown for reference only; its setting is the published Dream one.}
\label{tab:heldout_dream}
\small
\begin{tabular}{llccc}
\toprule
Held-out set & Probes & $\tau$ & Accuracy $\uparrow$ & NFE $\downarrow$ \\
\midrule
\multirow{10}{*}{GSM8K train (100)}
 & \textbf{20, inside main pass, masked only} & \textbf{0.95} & \textbf{74.0} & \textbf{42.2} \\
 & 20, inside main pass, masked only & 0.90 & 73.0 & 37.9 \\
 & 8, inside main pass, masked only & 0.95 & 71.0 & 42.8 \\
 & 12, inside main pass, masked only & 0.90 & 71.0 & 37.6 \\
 & 8, inside main pass, masked only & 0.90 & 70.0 & 38.2 \\
 & 12, inside main pass, masked only & 0.97 & 69.0 & 44.4 \\
 & 8, inside main pass, masked only & 0.97 & 68.0 & 45.2 \\
 & 20, inside main pass, masked only & 0.97 & 68.0 & 44.1 \\
 & 12, inside main pass, masked only & 0.95 & 66.0 & 42.1 \\
 & Mean-field (reference) & 0.95 & 78.0 & 68.6 \\
\midrule
\multirow{10}{*}{MBPP validation (90)}
 & \textbf{20, inside main pass, masked only} & \textbf{0.95} & \textbf{54.44} & \textbf{34.1} \\
 & 12, inside main pass, masked only & 0.97 & 54.44 & 34.9 \\
 & 20, inside main pass, masked only & 0.90 & 53.33 & 30.4 \\
 & 20, inside main pass, masked only & 0.97 & 53.33 & 35.5 \\
 & 12, inside main pass, masked only & 0.95 & 53.33 & 34.1 \\
 & 8, inside main pass, masked only & 0.97 & 52.22 & 35.2 \\
 & 8, inside main pass, masked only & 0.95 & 50.00 & 32.9 \\
 & 8, inside main pass, masked only & 0.90 & 48.89 & 30.3 \\
 & 12, inside main pass, masked only & 0.90 & 47.78 & 30.4 \\
 & Mean-field (reference) & 0.95 & 56.67 & 76.0 \\
\bottomrule
\end{tabular}
\end{table}

\end{document}